\documentclass[journal]{IEEEtran}
\usepackage{graphicx} 
\usepackage[utf8]{inputenc} 
\usepackage[T1]{fontenc}    
\usepackage{amsmath, amsfonts, amssymb}
\usepackage{booktabs}
\usepackage[numbers]{natbib}
\usepackage{lettrine}
\usepackage[ruled,vlined,linesnumbered]{algorithm2e}

\title{DSIP: A Dynamic Coordination Planner for Signal-Free Intersections Using Diffusion-Model-Based Multi-Agent Motion Planning}
\author{Qian Hu\thanks{Qian Hu and Songan Zhang are with the Global Institute of Future Technology, Shanghai Jiao Tong University, Shanghai 200240, China.}, Haoyang Peng\thanks{Haoyang Peng and Ming Yang are with the School of Automation and Intelligent Sensing, Shanghai Jiao Tong University, Shanghai 200240, China.}, Songan Zhang$^*$, Ming Yang, \textit{Member, IEEE}, and Hongtei Eric Tseng, \textit{Member, IEEE}\thanks{Hongtei Eric Tseng is with the Department of Electrical Engineering, The University of Texas at Arlington, Arlington, TX 76019 USA.}\thanks{\textit{Corresponding author: Songan Zhang (e-mail: songanz@sjtu.edu.cn)}.} }

\date{January 2026}

\begin{document}
\markboth{Journal of \LaTeX\ Class Files,~Vol.~14, No.~8, August~2021}%
{Shell \MakeLowercase{\textit{et al.}}: Bare Demo of IEEEtran.cls for IEEE Journals}

\maketitle

\begin{abstract}
Traffic signal control at urban intersections inherently introduces stop-and-go behavior, resulting in increased delays and reduced traffic efficiency, especially under high traffic demand. With the emergence of connected and automated vehicles (CAVs), trajectory-level coordination has emerged as a high-potential strategy to augment or transcend conventional phase-based management. This paper proposes DSIP (Diffusion-model-based Signal-free Intersection Planner), a multi-agent motion planning framework driven by a generative diffusion process. DSIP shifts the intersection management paradigm from discrete temporal phasing to continuous multi-vehicle trajectory optimization. This work evaluates the theoretical upper-bound performance of this coordination strategy under idealized communication and execution conditions to isolate the core benefits of the diffusion-driven approach. Using the SUMO platform, we evaluate DSIP across diverse four-leg intersection configurations. Experimental results demonstrate that DSIP significantly reduces average delay and maintains higher average speed compared to both fixed-time signal control and state-of-the-art reinforcement-learning-based controllers, particularly in medium- to high-density traffic. These findings suggest that diffusion-based trajectory planning provides a scalable and robust foundation for future autonomous intersection management. By unlocking latent intersection capacity through software-defined coordination, this approach offers a cost-effective pathway to improve urban traffic flow efficiency without requiring physical infrastructure expansion.
\end{abstract}
\begin{IEEEkeywords}
Signal-free intersection control, autonomous intersection management, multi-agent motion planning, diffusion-model-based planning.
\end{IEEEkeywords}

\section{Introduction}
\lettrine[lines=2, lraise=0.1, nindent=0em, findent=0.1em]{I}{n} 
the face of the global climate crisis, the transportation industry is undergoing a systemic transformation toward low-carbon and sustainable practices. A critical challenge in this transition is the environmental footprint of urban intersections, where idling and frequent stop-and-go maneuvers lead to incomplete fuel combustion, excessive energy waste, and elevated carbon emissions \cite{rakha2004}. Empirical research highlights the severity of this issue: when the average vehicle speed drops from 35 km/h to 15 km/h (i.e., traffic conditions deteriorate from normal to severely congested), the fuel consumption of gasoline, diesel, and liquefied petroleum gas (LPG) vehicles increases by 81 $\pm$ 16\%, 60 $\pm$ 15\%, and 107 $\pm$ 16\%, respectively, resulting in a significant increase in carbon emissions \cite{zhang2014}. Though less sensitive than conventional vehicles, electric vehicles (EVs) still consume 12--22\% more electricity during frequent stops and idling~\cite{wang2015}. This highlights the importance of reducing intersection delays for both conventional and electric vehicles. While numerous researchers have attempted to address this issue by optimizing driving strategies on individual vehicles, such as the eco-approach and departure (EAD) strategy \cite{ye2019}, such approaches still rely on driver operation and lack systemic coordination. Consequently, enhancing traffic flow at these critical nodes through automated control is imperative for achieving broader emission reduction goals.

Since the seminal Webster formula \cite{webster1958}, intersection control has evolved from pre-timed to actuated strategies, yet these remain unresponsive to dynamic or long-term traffic conditions \cite{yau2017}. Recently, reinforcement learning approaches employing complex reward functions \cite{wei2018} and hierarchical architectures \cite{tan2020} have achieved significant improvements; however, they still encounter the fundamental barrier of the curse of dimensionality \cite{yau2017}. Specifically, attempting to optimize phase sequences across multiple intersections or incorporating detailed vehicle states leads to an exponential expansion of the state-action space, hindering efficient policy convergence and limiting scalability under high-density traffic conditions.

Meanwhile, with the development of autonomous driving technology, especially connected and automated vehicles (CAVs) \cite{guanetti2018}, and the maturity of Vehicle-to-Everything (V2X) communication \cite{yoshizawa2023}, employing strategies such as virtual spring \cite{gong2024}, it becomes possible to control vehicles at signal-free intersections. Early pioneering works such as VICS~\cite{kamal2015} demonstrated the potential of trajectory-level coordination under a model predictive control framework, achieving significant improvements in delay and throughput compared to signalized control. Later, Malikopoulos et al.~\cite{malikopoulos2018} formulated a longitudinal speed optimization framework for energy-efficient and stop-free CAV passage at signal-free intersections. Lee and Park~\cite{lee2012} investigated cooperative vehicle intersection control under connected-vehicle environments and demonstrated the potential benefits of vehicle-level coordination compared with conventional signalized control. Xu et al.~\cite{xu2018} further studied distributed conflict-free cooperation for multiple connected vehicles at unsignalized intersections, reinforcing the feasibility of centralized and distributed coordination under idealized CAV assumptions. Despite their achievements, these conventional coordination methods struggle to characterize the complex, non-linear joint trajectory distributions of multiple interacting agents, often leading to sub-optimal or unsafe plans in dynamic environments.

Building upon this line of research, this paper investigates whether diffusion-model-based multi-agent motion planning can serve as an effective coordination mechanism for signal-free intersections. Unlike the aforementioned optimization-based approaches that rely on rigid, deterministic formulations, our diffusion-model-based method combines learned single-agent trajectory priors with a CBS-based coordination layer to resolve inter-vehicle conflicts. In this way, our study evaluates the potential of continuous vehicle-level planning as an alternative to conventional signalized intersection control in future CAV-intensive traffic environments. Similar assumptions have been adopted in prior studies on signal-free intersection control and CAV-based coordination, such as \cite{xu2018}. Our work follows two core standard assumptions of the field:

\textbf{1)} The traffic stream consists exclusively of CAVs governed by a centralized coordination framework, ensuring precise controllability of all vehicles within the intersection conflict zone;

\textbf{2)} The V2X communication environment is assumed to be perfect with negligible transmission latency and packet loss, thereby isolating the evaluation to the core planning capabilities and upper-bound performance of the proposed method.

\begin{figure*}[!t]
\centering
\includegraphics[width=\linewidth]{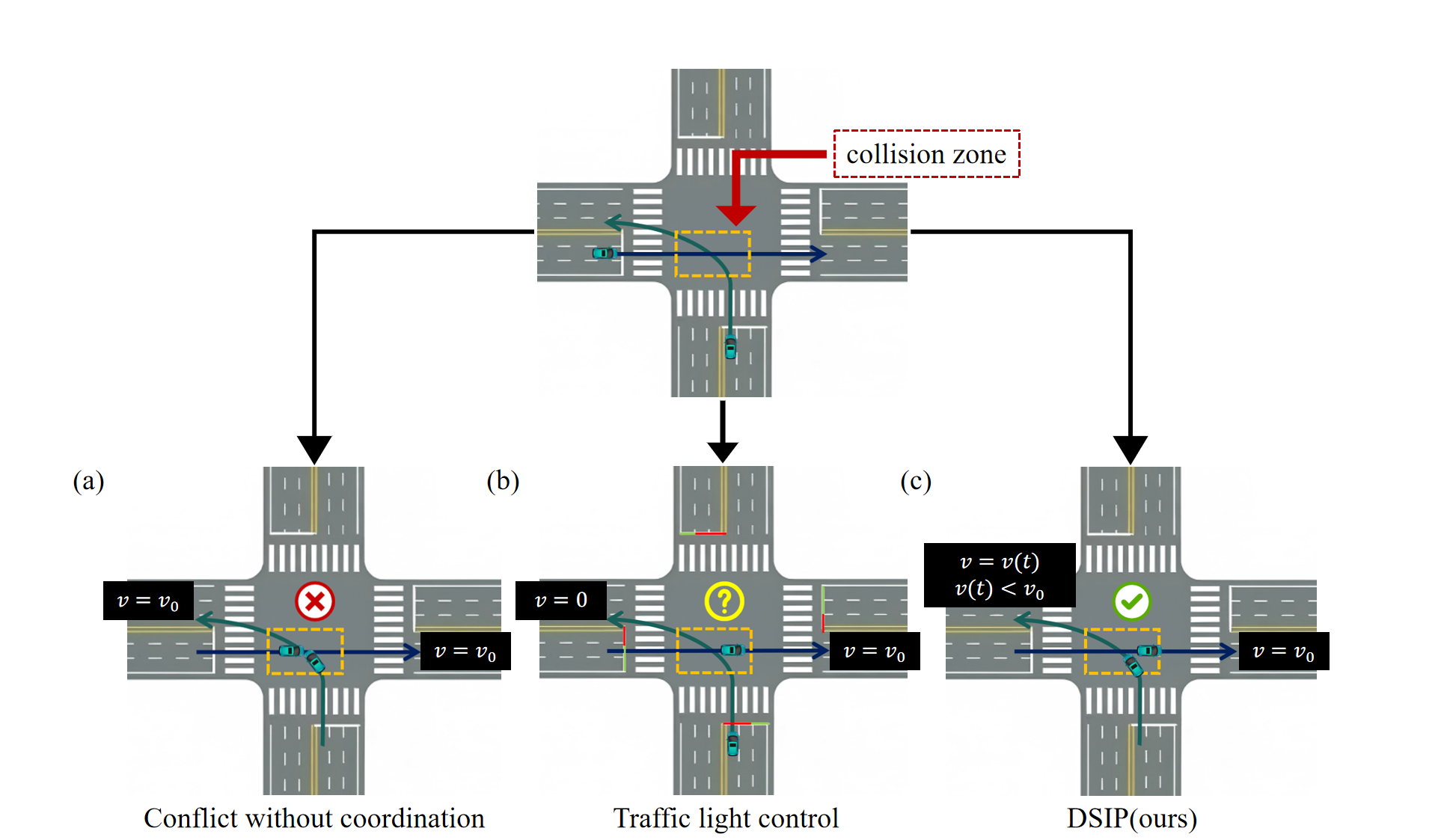} 
\caption{Inherent limitations of typical intersection management schemes and the core framework of this work: (a) Uncoordinated driving of connected and automated vehicles (CAVs) at signal-free intersections brings severe collision risk in the conflict zone; (b) Traditional fixed-cycle traffic light control eliminates collision risk, but introduces mandatory vehicle stops and large traffic delay, resulting in inherent low road utilization; (c) This paper proposes the DSIP scheme, which realizes collision-free and stop-free intersection passing through trajectory coordination of CAVs with fixed driving routes.}
\label{fig:demo}
\end{figure*}

In this paper, we introduce \textbf{DSIP (Diffusion-model-based Signal-free Intersection Planner)}, a diffusion-model-based multi-agent motion planner for urban signal-free intersections to solve the problem shown in Fig.~\ref{fig:demo}. Our main contributions are: 

\textbf{1)} We develop a diffusion-model-based multi-agent motion planner for signal-free intersection management, which jointly optimizes traffic efficiency for large-scale multi-vehicle coordination.

\textbf{2)} We propose a coordination strategy that combines pre-intersection trajectory initialization with periodic trajectory refresh and step-wise execution within the intersection, maintaining collision-free operation under the idealized SUMO execution setting.

\textbf{3)} We provide comprehensive simulation results with SUMO \cite{behrisch2011} demonstrating that trajectory-level coordination can achieve superior traffic efficiency and safety compared to traditional signalized control and state-of-the-art reinforcement-learning-based baselines.

The remainder of this paper is organized as follows. Section~\textrm{II} introduces some existing research findings related to the problem. The detailed settings of the simulation are included in Section~\textrm{III}. Section~\textrm{IV} describes the core methods we used, and briefly introduces the traffic signal settings used for the baseline. Sections~\textrm{V} and~\textrm{VI} present the simulation results and conclude the entire paper, respectively.

\section{Related Works}

To position our proposed DSIP framework within the existing literature, this section reviews relevant research across two major paradigms. We first examine Traffic Signal Control (Section II-A), which outlines the traditional and learning-based baselines for intersection management. We then delve into Multi-agent Motion Planning Methodologies (Section II-B), focusing on how CAVs leverage optimization and learning-driven approaches to achieve conflict-free coordination in traffic environments.

\subsection{Traffic Signal Control}

Intersection control has evolved from fixed-time methods \cite{webster1958} to adaptive systems like SCOOT \cite{hunt1981} and SCATS \cite{sims1980}, which leverage real-time data but primarily optimize local intersections. Subsequent arterial coordination \cite{robertson1969}, \cite{wong1996} expanded scope to corridors, often at the expense of side streets, while network-wide frameworks like PRODYN \cite{henry1984} achieved broader balance. However, these optimization-based control methods face the dual challenges of insufficient flexibility and high system complexity in practical deployment.

Driven by advancements in artificial intelligence and vehicle-infrastructure cooperation, deep learning and reinforcement learning (RL) have emerged as prominent approaches for signal control. Early foundational works demonstrated the feasibility of using Deep Q-Networks for adaptive timing \cite{wei2018,van2016,zhao2019}. For instance, Wei et al. \cite{wei2018} employed a DQN framework to learn optimal phase actions, while Tan et al. \cite{tan2020} introduced hierarchical architectures for network-wide coordination. To further address scalability, advanced frameworks such as MP-Light \cite{chen2020}, CoLight \cite{colight}, and AttentionLight \cite{attentionlight} were developed to enhance network-level cooperation and adaptivity. While these methods represent the state-of-the-art in intelligent signal control, they fundamentally operate within the phase-based timing paradigm. Consequently, they cannot fully eliminate the stop-and-go behavior inherent to signalized intersections, which motivates our exploration of signal-free trajectory-level coordination.

\subsection{Multi-Agent Motion Planning Methodologies}
Traditional multi-agent planning primarily relies on graph search, sampling-based algorithms, and Multi-Agent Path Finding (MAPF) formulations
\cite{hart1968,koenig2002,kavraki1996,lavalle1998,karaman2011,stern2019}, which inherently suffer from exponential computational growth in high-density scenarios \cite{sharon2015}. To overcome these scalability bottlenecks, data-driven methodologies have shifted the paradigm toward learned motion representations. While imitation learning \cite{pomerleau1988,abbeel2004,ho2016} and early generative frameworks (e.g., Motion Planning Diffusion \cite{carvalho2023}) successfully bypass explicit search to generate smooth trajectories, they are predominantly restricted to single-agent scenarios and lack inherent mechanisms for collective coordination.

To bridge this gap, recent advancements integrate data-driven approaches into multi-agent planning to address scalability challenges~\cite{stern2019}. While methods like MotionDiffuser~\cite{jiang2023} and Diffusion Policy~\cite{chi2025} excel in trajectory generation, they often struggle with explicit multi-agent coordination~\cite{sartoretti2019}. In contrast, Multi-agent Motion Planning with Diffusion Models (MMD)~\cite{shaoul2025} offers a scalable framework by hybridizing search strategies with generative models and soft constraints. However, direct deployment in structured urban intersections requires additional scenario-specific modeling for vehicle routes, road geometry, and traffic rules.

Synthesizing the aforementioned analysis, the diffusion-based multi-agent planning framework in~\cite{shaoul2025} is adopted as the methodological foundation of this study. This framework provides a generative formulation for coordinated motion planning, making it well suited for modeling continuous trajectory interactions among multiple agents. Compared with conventional search- or optimization-based planning methods, diffusion-based planning offers a flexible and scalable mechanism for generating diverse, feasible, and smooth trajectories under complex interaction constraints.

Building upon this foundation, this work establishes a signal-free intersection coordination framework tailored to CAV coordination in structured urban traffic environments. Specifically, the proposed method combines single-agent diffusion-based trajectory generation with search-based multi-vehicle conflict resolution, enabling trajectory-level coordination in dense vehicle flows. This generative formulation enables the output of socially compliant, collision-free trajectories without relying on discrete phase-based signal control. In this way, the proposed framework bridges diffusion-based multi-agent planning with signal-free intersection management, providing a robust basis for evaluating trajectory-level coordination in future CAV-intensive traffic systems.

\section{Simulation}

This section describes the SUMO-based simulation environment used to evaluate the proposed DSIP framework. To ensure a systematic comparison with traffic signal baselines, we design a series of controlled intersection scenarios by varying lane configurations and traffic densities. The simulation explicitly models intersection geometry, vehicle flows, and traffic rules. By isolating traffic demand and network topology, this setup facilitates a comprehensive assessment of efficiency, safety, and capacity across different operational conditions.

\subsection{Basic Settings of the Intersection for Experiment}  

Consistent with the idealized coordination assumptions outlined in Section I, the simulation environment standardizes intersection geometry and operational rules across all scenarios. Specifically, the evaluation is conducted across three distinct lane configurations and five traffic density levels, spanning from normal to highly saturated regimes. This controlled setup isolates the core algorithmic performance of the DSIP framework by eliminating confounding variables such as infrastructure heterogeneity and communication latency.

\subsubsection{Intersection Geometry and Traffic Demand}

In this work, we simulate three isolated four-leg intersections with varying lane configurations: two-way roads with 4, 6, and 8 lanes. These scenarios share identical geometric and operational parameters, differing only in lane count to isolate the impact of intersection scale. For each configuration, five distinct traffic density levels were defined, as summarized in Table~\ref{tabletrafficflow}, enabling systematic assessment across low- to high-saturation regimes.

\begin{table}[!t]
\caption{Experimental Traffic Flow Density Settings}
\label{tabletrafficflow}
\centering
\begin{tabular}{lcc}
\toprule
Direction & L\&R & S \\
\midrule
Minimum Density (MinD)  & 240 & 480  \\
Low Density (LowD)      & 300 & 600  \\
Medium Density (MedD)   & 384 & 768  \\
High Density (HigD)     & 480 & 960  \\
Maximum Density (MaxD)  & 600 & 1200 \\
\bottomrule
\addlinespace
\multicolumn{3}{l}{L/R refers to left-turning and right-turning vehicles;} \\
\multicolumn{3}{l}{S refers to straight-through vehicles;} \\
\multicolumn{3}{l}{The unit of traffic density is $\text{veh}/(\text{h}\cdot\text{lane})$.} \\
\end{tabular}
\end{table}

Here, the traffic demand is scaled proportionally with the number of lanes to maintain consistent saturation levels across different intersection configurations. The turning movement distribution follows a fixed ratio of $\text{Left} : \text{Straight} : \text{Right} = 1 : 2 : 1$, reflecting typical urban intersection patterns \cite{morris2010}. Regarding traffic density selection, the fourth and fifth density levels (HigD and MaxD in Table~\ref{tabletrafficflow}) represent high-saturation regimes that challenge conventional signalized control strategies. 

Vehicles depart from the four approaches of the intersection. The departure time and initial location are controlled by a random seed. A Gaussian departure process is used to introduce actual time variability while maintaining controllable average traffic demand. The departure interval follows a Gaussian distribution:

\begin{equation}
\Delta T_{\mathrm{interval}} \sim \mathcal{N}\!\left(\frac{3600}{nd},\,1\right),
\label{eq_density}
\end{equation}
where $n$ denotes the number of lanes, and $d$ is the traffic density value as specified in Table~\ref{tabletrafficflow}.

\subsubsection{Traffic Regulations and Safety Constraints}

The simulated intersection was modeled as an urban road segment with a centerline, and the maximum speed limit was set to 50 km/h (13.9 m/s) \cite{PRC_Traffic_Reg_2004}. To ensure operational safety, a minimum inter-vehicle gap of 1.5 meters was enforced; violations of this threshold were recorded as potential collisions \cite{GB51286}. All remaining vehicle dynamics parameters, including acceleration profiles and car-following behavior, were governed by SUMO's default microscopic traffic models to maintain consistency with established simulation practices.

\subsection{Metrics for Measuring Traffic Efficiency}

The primary metrics for assessing traffic efficiency center on average time loss and average speed, both of which serve as proxies for energy consumption and emission performance, aligning with the low-carbon motivation of this study. 

\subsubsection{Average Time Loss} 
Following SUMO's standard definition, the time loss for vehicle $i$ is calculated as the difference between its actual travel time and the theoretical optimal travel time under free-flow conditions:
\begin{equation}
\text{TimeLoss}_i = T_{\text{actual},i} - T_{\text{optimal},i} = T_{\text{actual},i} - \frac{d_i}{v_{\text{max},i}},
\end{equation}
where $T_{\text{actual},i}$ denotes the actual travel time of vehicle $i$ through the intersection region, $d_i$ represents the trip distance within the evaluation zone, and $v_{\text{max},i}$ is the maximum allowed speed on the corresponding road segment (set to 50 km/h in our experimental setup). At each logging step, the average time loss is computed over all vehicles currently active in the simulation:
\begin{equation}
\overline{\text{TimeLoss}} = \frac{1}{N} \sum_{i=1}^{N} \text{TimeLoss}_i.
\end{equation}
This metric quantifies the cumulative delay induced by intersection conflicts and control strategies, with higher values indicating greater efficiency losses and corresponding increases in fuel consumption due to prolonged idling and stop-and-go behavior.

\subsubsection{Average Speed} 
At each logging step, the average speed is computed as the mean instantaneous speed of all vehicles currently active in the simulation:
\begin{equation}
\bar{v} = \frac{1}{N_{\mathrm{act}}} \sum_{i=1}^{N_{\mathrm{act}}} v_i,
\end{equation}
where $N_{\mathrm{act}}$ denotes the number of active vehicles at the current logging step, and $v_i$ is the instantaneous speed of vehicle $i$ reported by SUMO. Alternatively, for individual vehicle analysis, the mean speed can be expressed as:
\begin{equation}
\bar{v}_i = \frac{1}{T_i} \int_{0}^{T_i} v_i(t) \, dt \approx \frac{1}{K} \sum_{k=1}^{K} v_i(t_k),
\end{equation}
where $v_i(t_k)$ represents the instantaneous speed of vehicle $i$ at simulation time step $t_k$, and $K$ is the total number of time steps during the traversal period. This metric reflects the smoothness of traffic flow, with higher and more stable speeds indicating reduced acceleration/deceleration events and lower energy waste.

We analyze the temporal evolution of average delay and compare steady-state speeds across control methods under identical intersection geometry and traffic demand. To evaluate operational safety, we additionally monitor collision avoidance performance by recording the count of critical proximity events (inter-vehicle gaps below the 1.5 m safety threshold) and abrupt deceleration incidents, defined as longitudinal deceleration exceeding SUMO's emergency braking threshold. These safety metrics provide a conservative assessment of the planner's robustness under high-density coordination.

\section{Methodology}

This section details the methodology of the proposed signal-free intersection management framework. We first describe the diffusion-model-based multi-agent motion planning algorithm designed for signal-free intersection. The formulation of motion optimization, collision avoidance constraints, and vehicle-level control logic is then presented to clarify how continuous, conflict-free coordination is achieved in real time. Finally, the configuration of the signalized baseline is briefly summarized.

\subsection{Diffusion-Model-Based Multi-Agent Motion Planning Algorithm}

To construct the proposed planner for signal-free intersections, we formulate the coordination task as a constrained trajectory generation problem in a two-dimensional road workspace. The SUMO road network is projected onto a global Cartesian coordinate system, where lane centerlines, lane shapes, lane boundaries, and map limits are obtained through the TraCI interface. For vehicle $i$, the route assigned by SUMO is denoted by $\Gamma_i$, which specifies its entry edge, turning movement, and exit edge. Based on this route, a start position and a destination position are extracted from the first and last road segments, respectively.

For vehicle $i$, the generated trajectory over a finite horizon is written as
\begin{equation}
\mathbf{X}_i
=
\{\mathbf{x}_{i,0},\mathbf{x}_{i,1},\ldots,\mathbf{x}_{i,H}\},
\label{eq_vehicle_traj_def}
\end{equation}
where $H$ is the planning horizon and $h$ is the discrete time index. The trajectory state is defined as
\begin{equation}
\mathbf{x}_{i,h}
=
[x_{i,h},y_{i,h}]^\top,
\label{eq_vehicle_state_def}
\end{equation}
where $(x_{i,h},y_{i,h})$ denotes the planned position of vehicle $i$ in the global SUMO map at time step $h$. The trajectory is later executed in SUMO by moving the vehicle to the corresponding planned positions, while the step-wise velocity is calculated from the displacement between adjacent trajectory points.

In this work, the diffusion planner is not trained from scratch for the intersection scenario. Instead, we employ a pre-trained single-agent diffusion planner following the framework in~\cite{shaoul2025}, and couple it with SUMO-specific route extraction, lane constraints, and multi-vehicle conflict resolution. The diffusion model provides a learned single-agent trajectory prior, while the CBS-based coordination layer handles conflicts among multiple vehicles. Let $\Theta_i$ denote the task condition of vehicle $i$, including its start position, destination position, assigned route $\Gamma_i$, planning horizon, and map validity constraints. Let $\mathcal{B}_i$ denote the constraint set imposed on vehicle $i$, including lane-following soft constraints and collision-avoidance constraints generated during multi-vehicle planning.

The desired trajectory is obtained by maximizing the posterior probability conditioned on the traffic task and constraints:
\begin{equation}
\mathbf{X}_i^{*}
=
\mathop{\arg\max}_{\mathbf{X}_i}
\log P_{\theta}(\mathbf{X}_i|\Theta_i,\mathcal{B}_i).
\label{eq_posterior_max}
\end{equation}
Equivalently, this posterior maximization can be written as a cost minimization problem:
\begin{equation}
\mathbf{X}_i^{*}
=
\mathop{\arg\min}_{\mathbf{X}_i}
\mathcal{L}_{i}^{\mathrm{post}}(\mathbf{X}_i),
\label{eq_posterior_min}
\end{equation}
where the posterior cost is defined as
\begin{equation}
\mathcal{L}_{i}^{\mathrm{post}}
=
\mathcal{L}_i(\mathbf{X}_i;\Theta_i,\mathcal{B}_i)
-
\lambda_{\mathrm{p}}\log P_{\theta}(\mathbf{X}_i).
\label{eq_dsip_objective}
\end{equation}
Here, $P_{\theta}(\mathbf{X}_i)$ denotes the trajectory prior represented by the pre-trained diffusion model, $\lambda_{\mathrm{p}}$ is the weight of this learned prior, and $\mathcal{L}_i$ represents the traffic-related trajectory cost. In the SUMO intersection scenario, this objective means that the generated trajectory should remain close to the learned smooth-motion distribution, satisfy the given start and destination conditions, follow the assigned route as much as possible, and avoid conflicts with other vehicles.

The traffic-related trajectory cost is written as
\begin{equation}
\mathcal{L}_i
=
\lambda_{\mathrm{g}}\mathcal{L}_{\mathrm{goal}}
+
\lambda_{\mathrm{l}}\mathcal{L}_{\mathrm{lane}}
+
\lambda_{\mathrm{s}}\mathcal{L}_{\mathrm{smooth}}
+
\lambda_{\mathrm{c}}\mathcal{L}_{\mathrm{col}},
\label{eq_total_loss}
\end{equation}
where $\mathcal{L}_{\mathrm{goal}}$ penalizes deviation from the destination position extracted from the SUMO route, $\mathcal{L}_{\mathrm{lane}}$ encourages the generated trajectory to remain near the lane points obtained from TraCI, $\mathcal{L}_{\mathrm{smooth}}$ suppresses abrupt trajectory changes, and $\mathcal{L}_{\mathrm{col}}$ penalizes potential conflicts with other vehicles. The weights $\lambda_{\mathrm{g}}$, $\lambda_{\mathrm{l}}$, $\lambda_{\mathrm{s}}$, and $\lambda_{\mathrm{c}}$ control the relative importance of destination reaching, lane following, smoothness, and collision avoidance.

The diffusion prior follows a standard denoising process in trajectory space. A clean single-agent trajectory $\mathbf{X}_i^{0}$ is progressively perturbed into a noisy trajectory $\mathbf{X}_i^{m}$:
\begin{equation}
q(\mathbf{X}_i^{m}|\mathbf{X}_i^{0})
=
\mathcal{N}
\left(
\sqrt{\bar{\alpha}_{m}}\mathbf{X}_i^{0},
(1-\bar{\alpha}_{m})\mathbf{I}
\right),
\label{eq_forward_process}
\end{equation}
where $m\in\{1,\ldots,M\}$ is the diffusion step, $M$ is the total number of diffusion steps, $\alpha_m=1-\beta_m$, $\bar{\alpha}_{m}=\prod_{\ell=1}^{m}\alpha_{\ell}$, and $\beta_m$ is the variance schedule. In the present framework, this process is provided by the pre-trained single-agent diffusion planner and is used as a generative prior for two-dimensional trajectory generation.

During online planning, the reverse denoising process starts from a noisy trajectory and gradually recovers a feasible trajectory. At each reverse step, the network predicts a denoised trajectory mean:
\begin{equation}
\boldsymbol{\mu}_{i}^{m-1}
=
\boldsymbol{\mu}_{\theta}
(\mathbf{X}_i^{m},m,\Theta_i),
\label{eq_denoised_mean}
\end{equation}
where $\boldsymbol{\mu}_{\theta}$ is the denoising network and $\Theta_i$ provides the start and destination conditions of vehicle $i$. The task-related guidance gradient is computed as
\begin{equation}
\mathbf{g}_{i}^{m-1}
=
\nabla_{\mathbf{X}_i}
\mathcal{L}_i
(
\boldsymbol{\mu}_{i}^{m-1};
\Theta_i,\mathcal{B}_i
).
\label{eq_guidance_gradient}
\end{equation}
The guided reverse sampling step is then expressed as
\begin{equation}
\mathbf{X}_i^{m-1}
\sim
\mathcal{N}
\left(
\widehat{\boldsymbol{\mu}}_{i}^{m-1},
\beta_{m-1}\mathbf{I}
\right),
\label{eq_reverse_sampling}
\end{equation}
where the guided mean is
\begin{equation}
\widehat{\boldsymbol{\mu}}_{i}^{m-1}
=
\boldsymbol{\mu}_{i}^{m-1}
-
\eta\beta_{m-1}\mathbf{g}_{i}^{m-1}.
\label{eq_reverse_guidance}
\end{equation}
Here, $\eta$ is the guidance step size. The negative gradient direction in Eq.~\eqref{eq_reverse_guidance} guides the sampled trajectory toward lower task cost, so that lane-following, smoothness, and collision-related constraints can influence the denoising process.

Lane information from SUMO is introduced as soft geometric constraints. For the current lane of vehicle $i$, a set of reference points is sampled from the lane shape:
\begin{equation}
\mathcal{P}_{i}^{\mathrm{lane}}
=
\{\mathbf{p}_{i,1},\mathbf{p}_{i,2},\ldots,\mathbf{p}_{i,N_i}\},
\label{eq_lane_points}
\end{equation}
where $\mathbf{p}_{i,n}\in\mathbb{R}^{2}$ is the $n$-th reference point on the lane centerline and $N_i$ is the number of sampled lane points. These points are associated with the planning horizon and added to the single-agent planner as soft constraints. The corresponding lane-following cost can be written as
\begin{equation}
\mathcal{L}_{\mathrm{lane}}
=
\sum_{h=0}^{H}
d^{2}
\left(
\mathbf{x}_{i,h},
\mathcal{P}_{i}^{\mathrm{lane}}
\right),
\label{eq_lane_cost}
\end{equation}
where $d(\mathbf{x}_{i,h},\mathcal{P}_{i}^{\mathrm{lane}})$ denotes the distance from the planned position $\mathbf{x}_{i,h}$ to the nearest lane reference point. This term does not force the trajectory to exactly coincide with the lane centerline, but encourages route-consistent motion within the SUMO road geometry.

For collision handling, each vehicle is represented by a planar position with a safety radius in the diffusion planning layer. A conflict-related constraint is denoted as
\begin{equation}
b
=
\left(
\mathbf{c}_{b},
\rho_b,
\mathcal{H}_{b}
\right),
\label{eq_constraint_element}
\end{equation}
where $\mathbf{c}_{b}$ is the center of a potential conflict region, $\rho_b$ is the safety radius, and $\mathcal{H}_{b}$ is the active time-index set of the constraint. The collision penalty for vehicle $i$ is written as
\begin{equation}
\mathcal{L}_{\mathrm{col}}
=
\sum_{b\in\mathcal{B}_i}
\sum_{h\in\mathcal{H}_{b}}
\left[
\rho_b
-
d_{i,b,h}
\right]_{+}^{2},
\label{eq_collision_loss}
\end{equation}
where $[z]_{+}=\max(z,0)$, and $d_{i,b,h}$ denotes the distance between vehicle $i$ and the conflict center at time step $h$:
\begin{equation}
d_{i,b,h}
=
\left\|
\mathbf{x}_{i,h}
-
\mathbf{c}_{b}
\right\|_{2}.
\label{eq_conflict_distance}
\end{equation}
This formulation matches the disk-based two-dimensional planning model used by the pre-trained planner and provides a conservative safety margin for SUMO vehicle coordination.

Multi-vehicle coordination is handled by a CBS-based conflict-resolution module. For the set of vehicles currently active in SUMO, each vehicle is assigned an individual diffusion planner with its own start and destination positions. The multi-vehicle trajectory set at a planning query is denoted as
\begin{equation}
\mathcal{X}
=
\{\mathbf{X}_{1},\mathbf{X}_{2},\ldots,\mathbf{X}_{n}\},
\label{eq_multi_vehicle_traj_set}
\end{equation}
where $n$ is the number of vehicles included in the current planning call. The CBS-based module checks conflicts among the generated trajectories and adds corresponding constraints to the affected low-level diffusion planners. If a conflict between vehicles $i$ and $j$ is detected around time step $h_c$, a new constraint can be represented as
\begin{equation}
b_{ij,h_c}
=
\left(
\mathbf{c}_{ij,h_c},
\rho_{ij},
[h_c-\Delta h,h_c+\Delta h]
\right),
\label{eq_new_constraint}
\end{equation}
where $\mathbf{c}_{ij,h_c}$ is the conflict center, $\rho_{ij}$ is the required safety radius, and $\Delta h$ defines the temporal width of the constraint. The affected single-agent planner then regenerates a trajectory under the updated constraint set through the guided denoising process in Eqs.~\eqref{eq_denoised_mean}--\eqref{eq_reverse_guidance}. In this way, the framework combines the generative ability of the pre-trained diffusion planner with search-based multi-vehicle conflict resolution.

For clarity, the above CBS-based multi-vehicle planning procedure is denoted as \textsc{CBS-Diffusion-Plan} in Algorithm~\ref{alg:dsip}. 
It takes the active single-agent diffusion planners as input, detects trajectory conflicts, 
adds the corresponding spatio-temporal constraints to the affected planners, 
and returns the updated multi-vehicle trajectory set.

Finally, the generated trajectory is converted into SUMO vehicle motion through TraCI. At each simulation step, the planned position is obtained from the stored trajectory, adjusted to remain within the map boundary when necessary, and sent to SUMO. The vehicle heading is calculated from two adjacent planned positions. Thus, the diffusion planner determines the desired position sequence, while SUMO provides the road network, vehicle route, and traffic-state update.

\subsection{Control Logic for Vehicle Motion and Trajectory }

To meet the real-time requirements of traffic scenarios, this paper adopts a two-stage coordination strategy under the fixed-route constraint of SUMO. The diffusion planner generates multi-vehicle trajectories periodically, while SUMO executes the stored trajectory points at each simulation step.

\subsubsection{Pre-calculation (Pre-intersection Stage)}

Before a vehicle crosses the stop line, SUMO's built-in lane-changing and routing information~\cite{erdmann2015} is used to determine its legal route. The proposed framework extracts the start and destination positions from the first and last edges of this route, checks whether they are within the map boundary, and initializes an individual diffusion planner for the vehicle. Lane reference points are also sampled from the current lane shape and added as soft constraints to encourage route-consistent motion.

A \textit{trajectory validity check} and multi-vehicle planning query are performed every four simulation steps. Since the SUMO simulation step is $0.1\,\text{s}$, the planning result is refreshed every $0.4\,\text{s}$. At each planning query, active vehicles are collected, invalid or departed vehicles are removed, and the CBS-based module is called to generate conflict-free trajectories for the valid vehicles. If some vehicles have invalid start or destination positions, they are excluded from the current planning query, while valid vehicles continue to be planned.

\subsubsection{Real-time fine-tuning (In-intersection Stage)}

After vehicles enter the interaction area of the intersection, the system continues to execute the latest planned trajectories at every SUMO simulation step. Although the vehicle position is updated every $0.1\,\text{s}$, the diffusion-based multi-vehicle planning result is refreshed every four steps to balance planning latency and traffic responsiveness. Newly active vehicles are initialized when they appear in SUMO, and the next planning query incorporates them into the multi-vehicle coordination process.

At each execution step, the target position is read from the stored trajectory. If the target position exceeds the map boundary, it is clipped to the valid map range. The vehicle heading is calculated from the direction between the current planned position and the next planned position. The vehicle is then moved to the target position through the TraCI interface, while keeping its SUMO route as the reference. This process enables the generated trajectory to be reflected in the microscopic traffic simulation without claiming direct low-level vehicle actuation.

\subsubsection{Path Confirmation (Legality \& Safety Validation)}

To ensure the legality and safety of all generated trajectories, a two-stage path confirmation mechanism is applied throughout the planning process:

\paragraph{Static path confirmation before vehicle departure}
The predefined SUMO route is checked before planner initialization. The start and destination positions extracted from the route must lie within the calculated map boundary. If a position is outside the valid range, it is clipped to the nearest valid map position. Lane polygons and lane shapes are used to verify whether the route-related positions are consistent with the road geometry.

\paragraph{Dynamic path confirmation during operation}
During simulation, the set of active vehicles is updated at every step. Vehicles that have departed are removed from the planner and trajectory records. Before each planning query, the start and destination positions of active vehicles are checked again. Only vehicles with valid planning states are included in the CBS-based multi-vehicle planning process, and only generated trajectories that pass the boundary and conflict checks are executed.

In the SUMO microscopic traffic simulation, the time step is fixed at $0.1\,\text{s}$, which is widely used for urban traffic flow modeling~\cite{behrisch2011}. The planned trajectory is executed as a sequence of positions. Therefore, the step-wise speed can be approximated by the displacement between two adjacent planned points:
\begin{equation}
\bar{v}
=
\frac{\sqrt{(\Delta x)^2+(\Delta y)^2}}{\Delta t},
\label{eq_speed_calc}
\end{equation}
where $\bar{v}$ denotes the average speed within one simulation step, $\Delta x$ and $\Delta y$ are the coordinate differences between adjacent planned positions, and $\Delta t=0.1\,\text{s}$ is the simulation step size.

When a potential conflict is detected in the current multi-vehicle planning query, the CBS-based module introduces additional constraints for the affected single-agent diffusion planners. The constrained planner then updates its trajectory through guided denoising:
\begin{equation}
\mathbf{X}_i^{m-1}
\sim
\mathcal{N}
\left(
\widehat{\boldsymbol{\mu}}_{i}^{m-1},
\beta_{m-1}\mathbf{I}
\right),
\label{eq_guided_sampling_control}
\end{equation}
where the guided mean is
\begin{equation}
\widehat{\boldsymbol{\mu}}_{i}^{m-1}
=
\boldsymbol{\mu}_{i}^{m-1}
-
\eta\beta_{m-1}
\nabla_{\mathbf{X}_i}
\mathcal{L}_i(\boldsymbol{\mu}_{i}^{m-1}).
\label{eq_guided_mean_control}
\end{equation}
Here, $\boldsymbol{\mu}_{i}^{m-1}$ is the denoised trajectory mean predicted by the diffusion planner, and $\nabla_{\mathbf{X}_i}\mathcal{L}_i$ includes lane-following, smoothness, and collision-related costs. This guidance term steers the sampled trajectory toward lower conflict risk and smoother motion before the trajectory is executed in SUMO. Algorithm~\ref{alg:dsip} summarizes the overall DSIP coordination process, including SUMO-based vehicle initialization, periodic CBS-based diffusion planning, and step-wise trajectory execution.

\begin{algorithm}[!t]
\caption{DSIP coordination process for signal-free intersections}
\label{alg:dsip}
\KwIn{
SUMO road network and routes; pre-trained single-agent diffusion planner $f_{\theta}$;
planning period $K_{\mathrm{p}}=4$; simulation step $\Delta t=0.1\,\mathrm{s}$;
maximum simulation step $T$
}
\KwOut{
Executed vehicle trajectories and logged traffic-efficiency metrics
}

Initialize TraCI and start the SUMO simulation\;
Extract lane shapes, lane boundaries, edge--lane mappings, and map bounds from SUMO\;
Initialize planner dictionary $\mathcal{F}$ and trajectory dictionary $\mathcal{X}$\;

\For{$t=0$ \KwTo $T-1$}{
    Advance SUMO by one simulation step\;
    Obtain the active vehicle set $\mathcal{V}_{t}$ from SUMO\;
    Remove departed vehicles from $\mathcal{F}$ and $\mathcal{X}$\;

    \For{each new vehicle $i\in\mathcal{V}_{t}$}{
        Obtain its SUMO route $\Gamma_i$\;
        Extract start position $\mathbf{s}_i$ and destination position $\mathbf{g}_i$
        from the first and last edges of $\Gamma_i$\;
        Clip $\mathbf{s}_i$ and $\mathbf{g}_i$ to the valid map range if necessary\;
        Sample lane reference points $\mathcal{P}_{i}^{\mathrm{lane}}$ from the current lane shape\;
        Initialize a single-agent diffusion planner
        $F_i \leftarrow f_{\theta}(\mathbf{s}_i,\mathbf{g}_i,\mathcal{P}_{i}^{\mathrm{lane}})$\;
        Store $F_i$ in $\mathcal{F}$\;
    }

    \If{$t \bmod K_{\mathrm{p}}=0$}{
        Select vehicles with valid start and destination states:
        $\mathcal{V}_{t}^{\mathrm{valid}}\subseteq\mathcal{V}_{t}$\;
        Generate multi-vehicle trajectories by CBS-based conflict resolution:
        $\{\mathbf{X}_i\}_{i\in\mathcal{V}_{t}^{\mathrm{valid}}}
        \leftarrow
        \textsc{CBS-Diffusion-Plan}(\{F_i\}_{i\in\mathcal{V}_{t}^{\mathrm{valid}}})$\;
        Store the planned trajectories in $\mathcal{X}$\;
    }

    \For{each vehicle $i$ with a stored trajectory $\mathbf{X}_i\in\mathcal{X}$}{
        Read the target position $\mathbf{x}_{i,h}$ from the stored trajectory\;
        Clip $\mathbf{x}_{i,h}$ to the valid map range if necessary\;
        Estimate the heading angle from $\mathbf{x}_{i,h+1}-\mathbf{x}_{i,h}$\;
        Move vehicle $i$ to $\mathbf{x}_{i,h}$ through the TraCI interface while keeping its SUMO route as reference\;
    }

    Log average time loss, average speed, and active vehicle count at the specified logging interval\;
}
\end{algorithm}

\subsection{Experiment Setup and Baselines}
For each traffic demand level at each intersection configuration, we evaluated five control strategies: the proposed diffusion-based multi-agent planner, traditional fixed-time signal control, and three representative reinforcement-learning-based baselines (MP-Light, CoLight, and AttentionLight). This experimental design yielded 3 intersections $\times$ 5 density levels $\times$ 5 methods = 75 distinct test configurations. For each configuration, we conducted 30 independent simulation runs with different random seeds, each with a simulation horizon of 1000 seconds, to ensure statistical robustness.

The traditional traffic signal scheme follows the traffic regulations in \cite{PRC_Traffic_Reg_2004} and adopts a four-phase signal scheme for right-hand traffic. The specific phase sequence within a single cycle is presented in Table~\ref{tabgreentimeallocation}. The calculations in this study are mainly based on Webster's formula \cite{webster1958} and the revised methods provided by the Highway Capacity Manual (HCM) \cite{morris2010}. The green light duration for each signal phase is determined by traffic volume and intersection scale. According to the traffic flow settings in Table~\ref{tabletrafficflow}, the green time allocated for left-turn movements is half of that for through traffic.

\begin{table}[!t]
  \centering
  \caption{Traffic Light Green Time Allocation} 
  \begin{tabular}{cccc} 
    \toprule
    Direction & Left Turn & Straight & Right Turn \\ 
    \midrule
    Green Time (s) & 20 & 40 & Always Green \\ 
    \bottomrule
  \end{tabular}
  \label{tabgreentimeallocation} 
\end{table}

For these learning-based baselines, we used their existing configurations and evaluated them under the same intersection geometries, traffic demand settings, simulation horizon, and random-seed protocol as DSIP. No additional task-specific tuning was introduced beyond the scenario interface required for SUMO-based evaluation. In contrast, many pioneering AIM methods lack publicly available code or require substantial engineering effort to interface with SUMO's TraCI protocol, making direct comparison challenging without introducing implementation-specific artifacts. Moreover, the core research question of this work is whether traffic signals remain necessary in CAV-intensive environments, which is directly addressed by benchmarking against the currently deployed standard and its most advanced learning-based variants. Comparisons with other laboratory-scale AIM prototypes, while valuable for algorithm-level benchmarking, are orthogonal to this paradigm-level evaluation and are prioritized in our future work.

MP-Light is a typical decentralized multi-agent reinforcement learning (MARL) framework. It decomposes traffic signal control into decentralized intersection-level or movement-level decisions, thereby reducing computational costs under low to medium traffic density. CoLight adopts a grid-based, regionally collaborative and centralized MARL framework. It aggregates subregional states to optimize the global policy. 
AttentionLight enhances state representation by embedding a multi-head attention mechanism into centralized MARL, assigning adaptive weights to critical traffic features such as vehicle speed and lane priority to improve adaptability to complex traffic scenarios. It outperforms earlier methods in mixed traffic flows.

For intersections with traffic lights, the intersection policy is \texttt{traffic-light}; for intersections without traffic lights, the intersection policy is \texttt{unregulated} \cite{PlainXMLSUMODocumentation}.

\section{Results and Discussion}

This section presents a systematic evaluation of the proposed DSIP framework under diverse traffic densities and intersection configurations. The analysis prioritizes three key performance indicators: traffic efficiency, operational robustness, and safety compliance. Initially, the proposed signal-free paradigm is benchmarked against traditional signalized control and state-of-the-art reinforcement-learning-based baselines to establish a performance baseline. Subsequently, performance trends are examined across varying density regimes, with specific emphasis on high-saturation and extreme traffic conditions to stress-test the coordination logic. 

\subsection{Performance Comparison among Different Methods}

This section presents a systematic evaluation of the proposed DSIP framework against conventional signalized control paradigms. The analysis prioritizes three key performance indicators: traffic efficiency (quantified by average time loss), operational stability (measured via average speed), and safety compliance (assessed through collision and emergency braking events). Special emphasis is placed on high-saturation scenarios, where the limitations of phase-based control become most pronounced. The following subsections detail the performance disparities observed in 75 distinct experimental configurations, providing empirical evidence for the viability of diffusion-based multi-agent coordination as a sustainable alternative to traffic signals.

\subsubsection{Average Time Loss}

\begin{figure}[ht]
\centering
\includegraphics[width=3.5in]{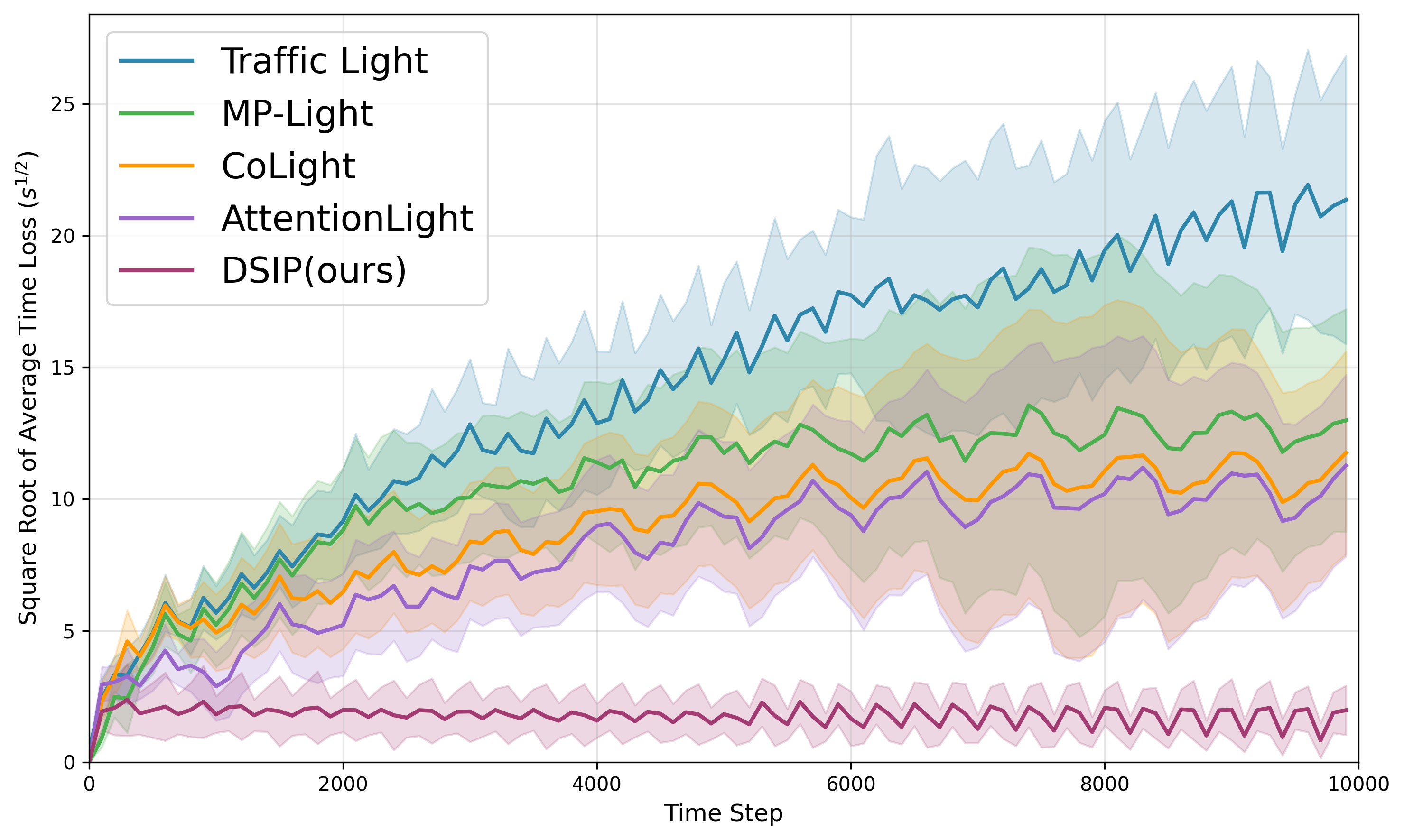} 
\caption{Average time loss under medium traffic density on a two-way four-lane road.}
\label{fig:tl2}
\end{figure}

\begin{figure}[ht]
\centering
\includegraphics[width=3.5in]{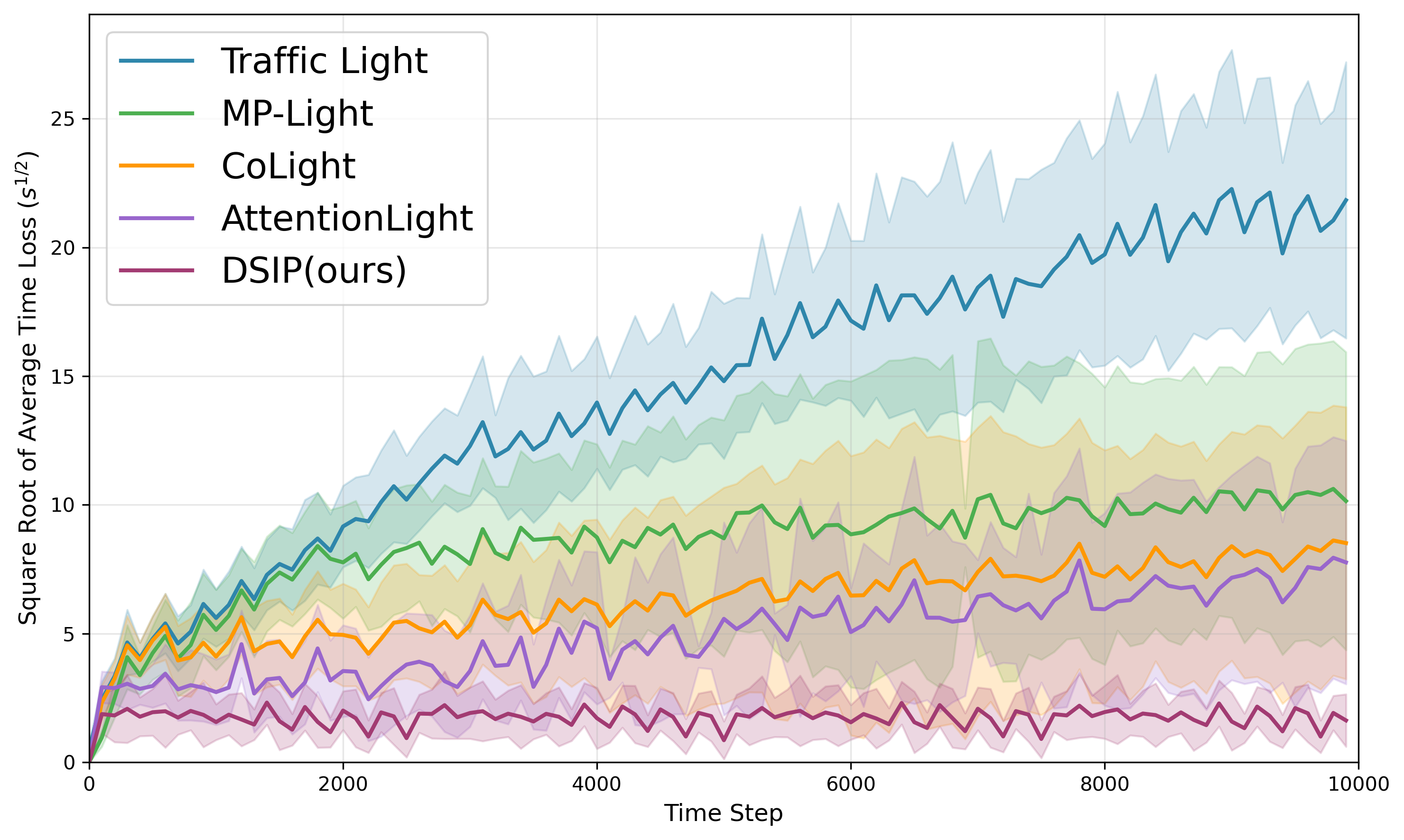} 
\caption{Average time loss under medium traffic density on a two-way six-lane road.}
\label{fig:tl3}
\end{figure}

\begin{figure}[ht]
\centering
\includegraphics[width=3.5in]{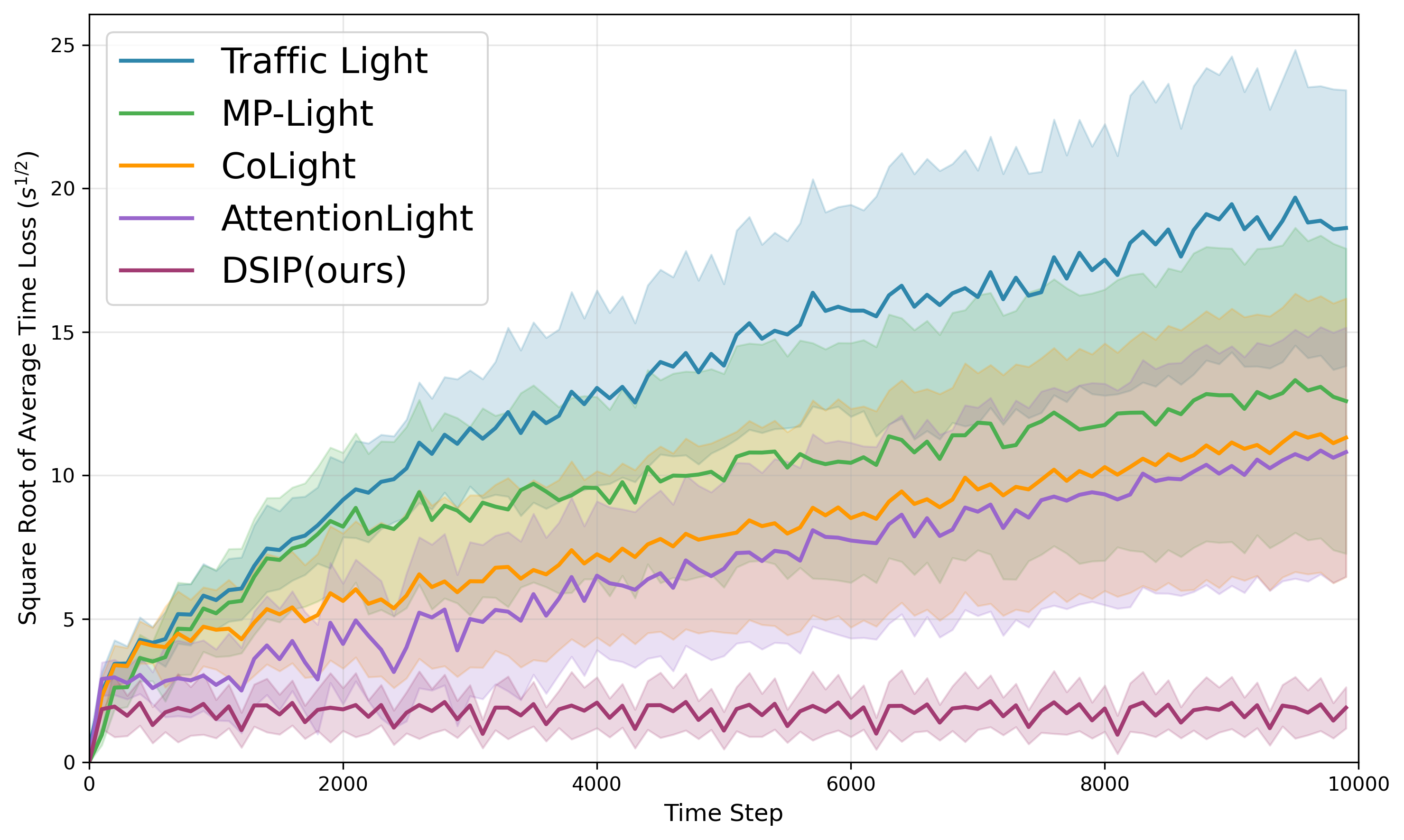} 
\caption{Average time loss under medium traffic density on a two-way eight-lane road.}
\label{fig:tl4}
\end{figure}

Taking the medium traffic density scenario (MedD in Table~\ref{tabletrafficflow}) as a representative case, we evaluated the temporal evolution of average time loss with a sampling interval of 10 seconds. For each control method under identical intersection geometry and traffic demand, results were aggregated over 30 independent simulation runs with distinct random seeds to ensure statistical reliability. The comparative performance across different lane configurations is presented in Figs.~\ref{fig:tl2}--\ref{fig:tl4}. Average time loss values were square-root transformed to compress the vertical axis, preventing low-delay methods from clustering and enabling clearer visual comparison.

Under medium traffic density, the proposed DSIP framework maintains average vehicle delay below 15 seconds throughout the simulation horizon, outperforming both traditional fixed-time signal control and state-of-the-art reinforcement-learning baselines. As depicted in Figs.~\ref{fig:tl2}--\ref{fig:tl4}, while signalized methods exhibit a monotonic increase in time loss indicative of queue accumulation, DSIP converges to a stable steady-state within the initial 2000 simulation steps. Furthermore, DSIP exhibits significantly narrower confidence intervals across 30 independent runs, indicating lower variance and higher robustness to stochastic demand fluctuations. In contrast, signalized baselines show widening variance over time, reflecting congestion-induced performance degradation. This stability underscores the advantage of signal-free trajectory coordination, which facilitates smooth vehicle passage through explicit vehicle-level conflict resolution rather than discrete phase segregation.

\subsubsection{Average Speed}

\begin{figure}[ht]
    \centering 
    \includegraphics[width=3.5in]{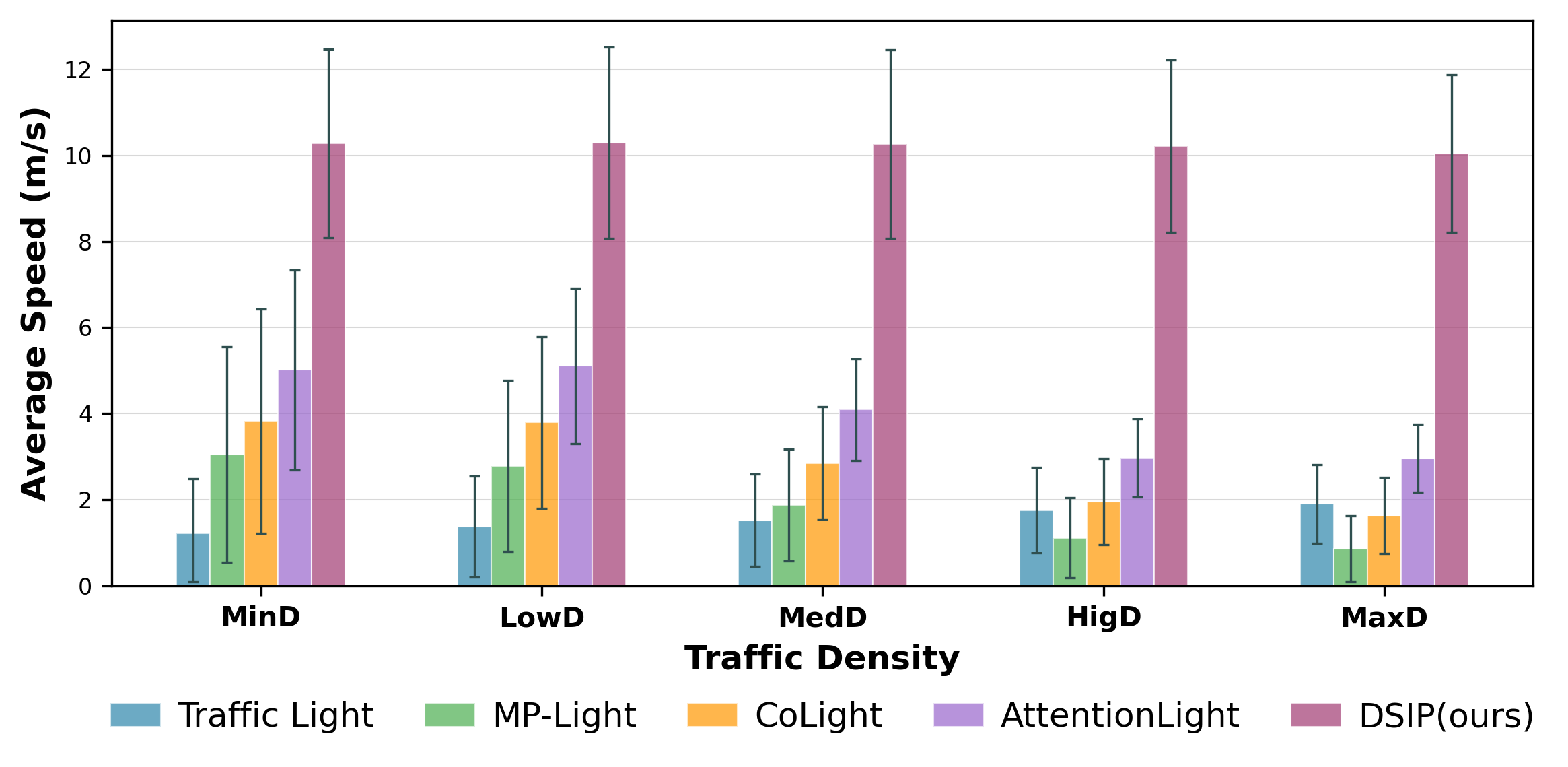} 
    \caption{Average speed under various traffic densities on a two-way four-lane road.}
    \label{fig:as2}
\end{figure}

\begin{figure}[ht]
    \centering 
    \includegraphics[width=3.5in]{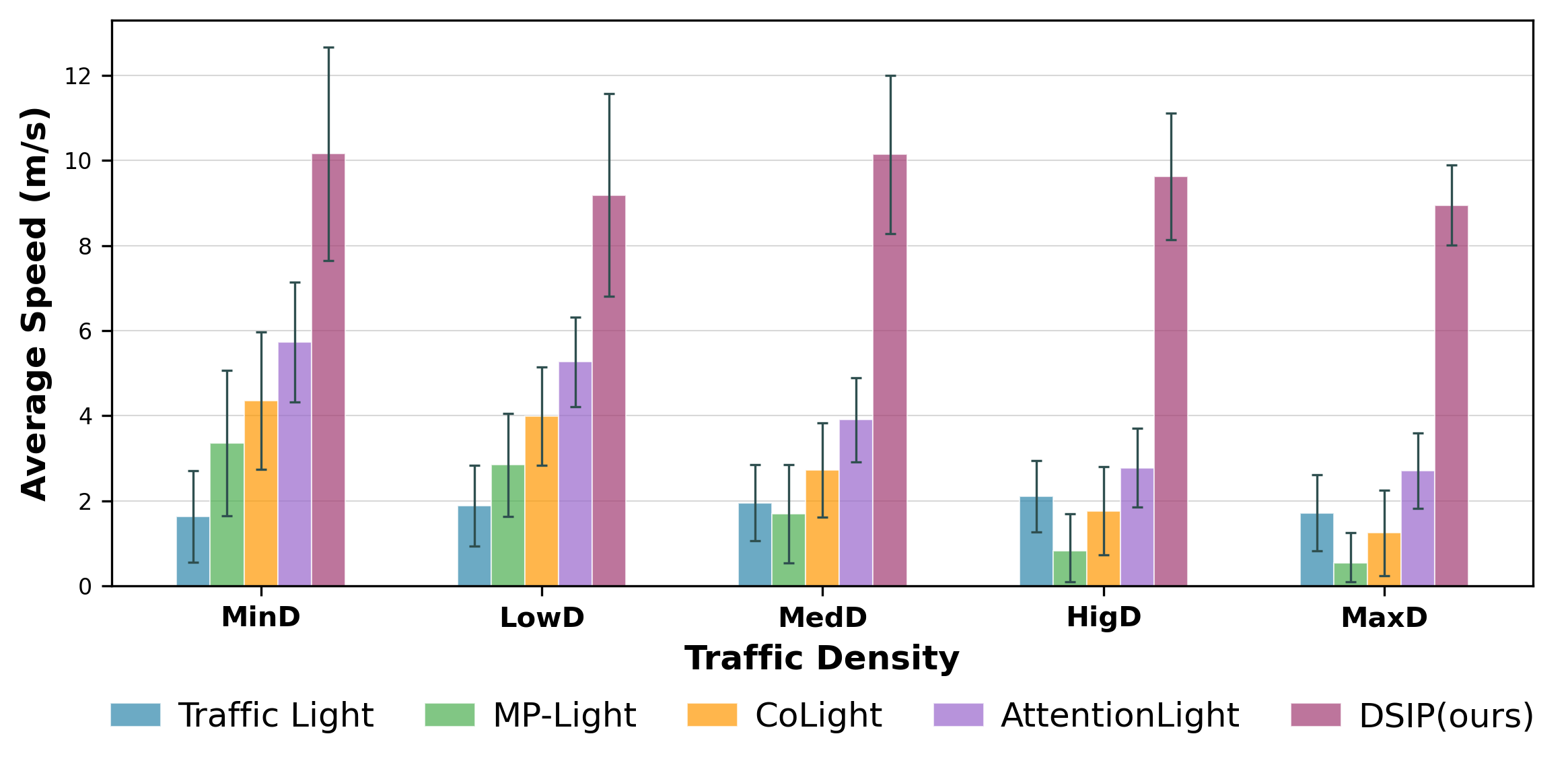} 
    \caption{Average speed under various traffic densities on a two-way six-lane road.}
    \label{fig:as3}
\end{figure}

\begin{figure}[ht]
    \centering 
    \includegraphics[width=3.5in]{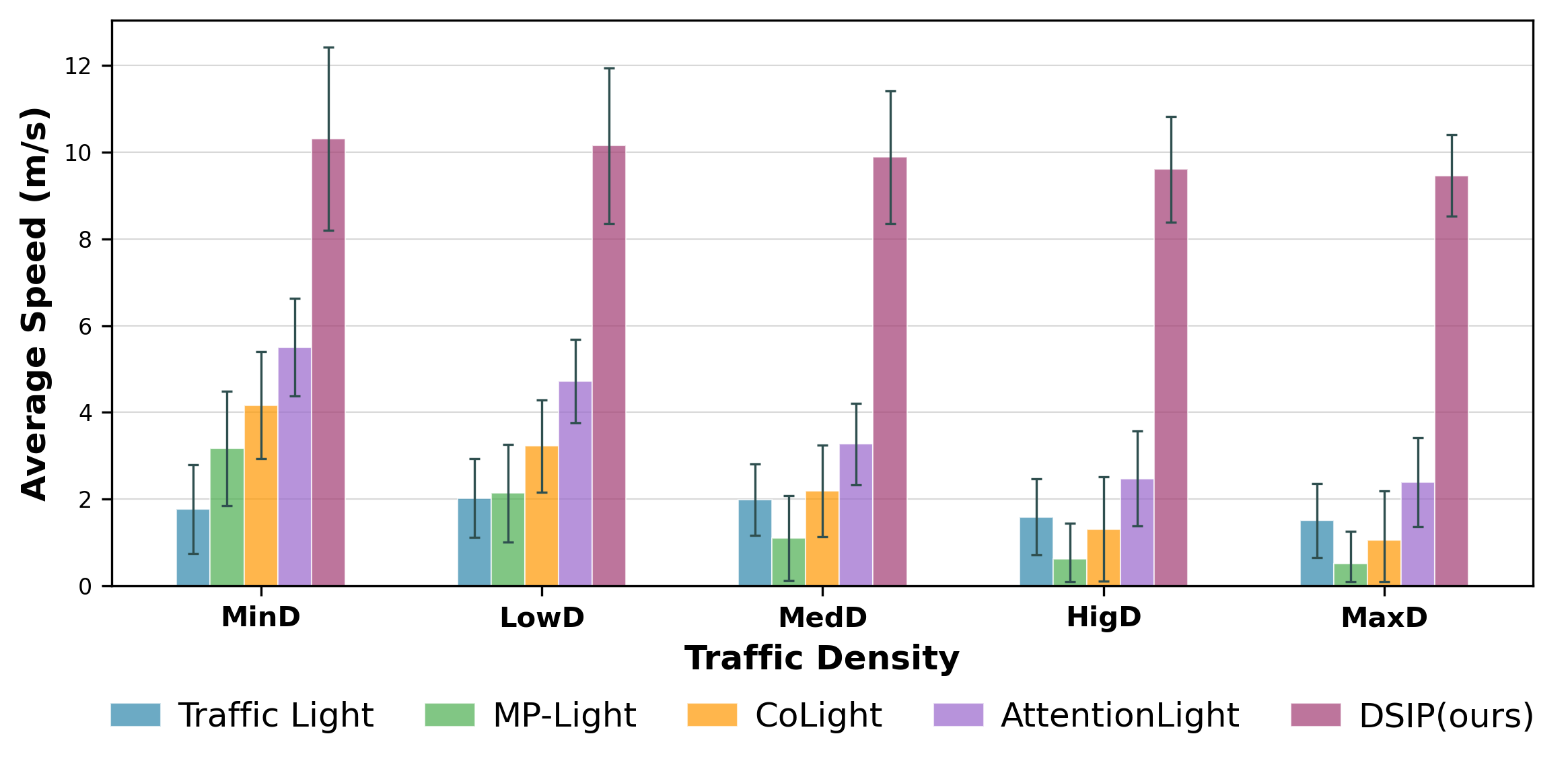} 
    \caption{Average speed under various traffic densities on a two-way eight-lane road.}
    \label{fig:as4}
\end{figure}

For each combination of intersection configuration, traffic density level, and control method, we conducted 30 independent simulation runs with distinct random seeds to ensure statistical reliability. The aggregated results, presented in Figs.~\ref{fig:as2}--\ref{fig:as4}, illustrate the distribution of average speeds across varying operational conditions. This multi-run evaluation protocol enables robust comparison of central tendency and variance, facilitating assessment of both efficiency gains and operational stability under the proposed signal-free paradigm versus conventional signalized baselines.

Figs.~\ref{fig:as2}--\ref{fig:as4} illustrate the average traversal speed across varying traffic densities and lane configurations. The proposed DSIP framework consistently achieves the highest average speed, maintaining approximately $9$--$10\,\mathrm{m/s}$ even under maximum density (MaxD). In contrast, all signal-based baselines exhibit significant performance degradation as density increases, with average speeds dropping below 3 m/s at MaxD. Notably, the performance gap between DSIP and baseline methods widens progressively from MinD to MaxD. Among baselines, MP-Light shows unstable performance at high densities, occasionally underperforming traditional fixed-time control. This superiority of DSIP in maintaining high-speed flow is consistent across two-way four-, six-, and eight-lane scenarios.

\subsubsection{Collision Avoidance}

Across all 450 DSIP simulation runs covering three intersection configurations and five traffic density levels, no collision warnings were observed under the idealized SUMO execution setting. Based on the experimental demand settings and simulation horizon, these runs correspond to over 2 million planned vehicle passages through the intersection. Only three instances of abrupt deceleration events, defined as longitudinal deceleration exceeding SUMO's emergency braking threshold, were observed throughout the entire evaluation. This near-zero incident rate demonstrates that trajectory-level coordination via diffusion-based planning can maintain safe multi-agent operation with reasonable safety margins under complex, high-saturation traffic scenarios, supporting the viability of the signal-free paradigm under the idealized simulation setting.

\subsection{Traffic Evolution Analysis}

\begin{figure*}[!t]
    \centering 
    \includegraphics[width=\linewidth]{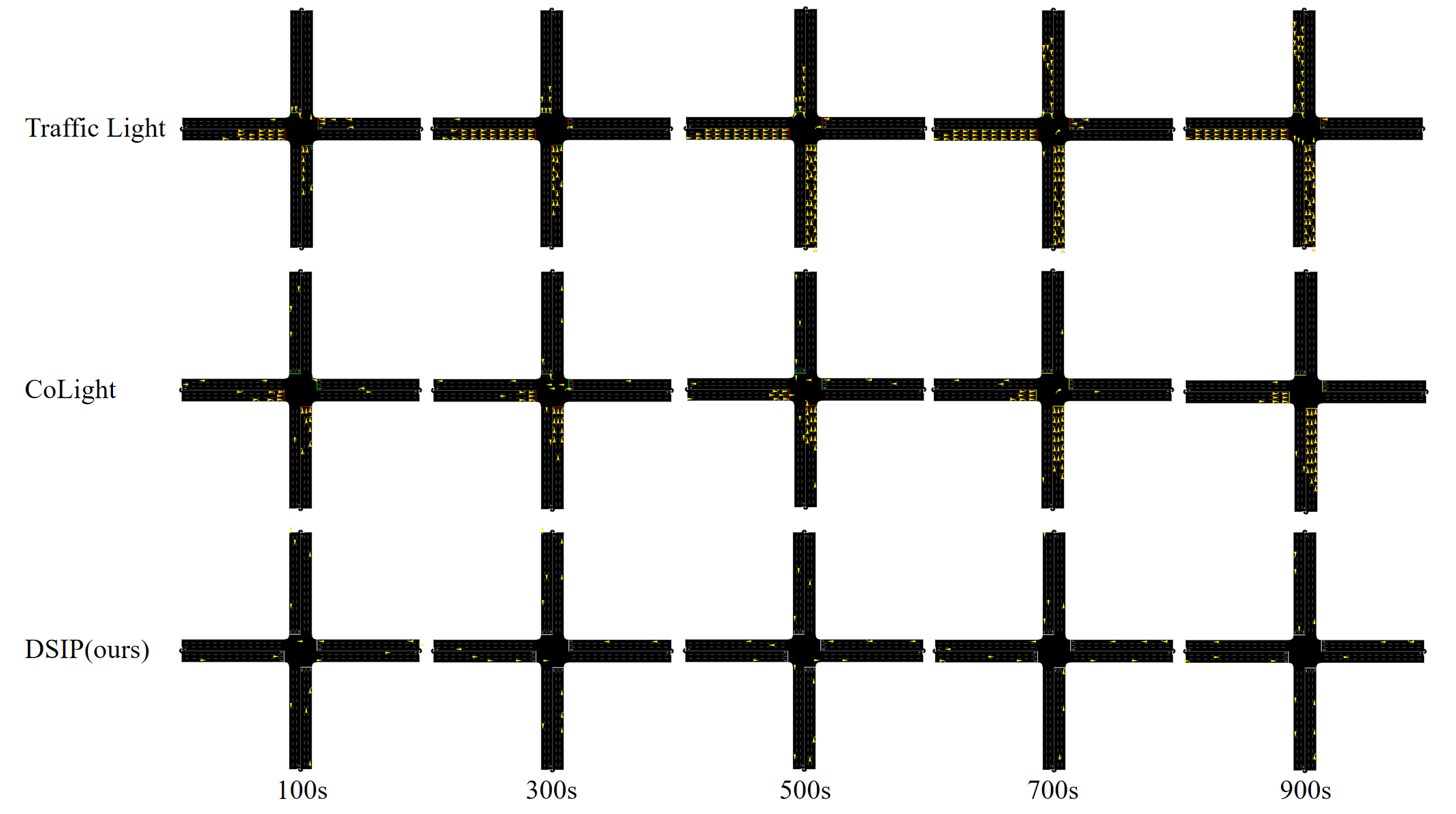} 
    \caption{Temporal evolution of traffic states under different control strategies at a four-leg intersection. The experimental scenario shown in the snapshots is a two-way six-lane intersection under medium density (MedD). From top to bottom: pre-timed signal control, CoLight-based adaptive signal control, and the proposed DSIP method. Snapshots are taken at different simulation time steps under identical traffic demand and random seed settings. Signal-based approaches exhibit evident queue accumulation due to phase constraints, while CoLight improves early-stage efficiency but still suffers from congestion buildup in later stages. In contrast, the proposed method maintains smoother traffic flow and avoids long queue formation through continuous multi-agent trajectory coordination.}
    \label{fig:evolution}
\end{figure*}

To further illustrate the dynamic differences among control strategies, Fig.~\ref{fig:evolution} presents a representative visualization of traffic evolution under identical experimental settings. We select CoLight as a representative RL-based baseline due to the shared phase-based control paradigm among reinforcement learning approaches.

As shown in the figure, fixed-time signal control leads to rapid queue accumulation caused by periodic stopping, while CoLight alleviates congestion in the early stage but gradually exhibits queue buildup under high traffic demand. In contrast, the proposed DSIP method achieves a more stable traffic state, where vehicle movements remain continuous and large-scale congestion is effectively mitigated.

These observations are consistent with the quantitative results presented earlier, and further demonstrate that multi-agent coordination can effectively reduce congestion compared to phase-based control strategies.

\subsection{Discussion}
Collectively, the results across Figs.~\ref{fig:tl2}--\ref{fig:as4} reveal a consistent performance hierarchy, where DSIP maintains superior efficiency and stability across all lane configurations and density regimes compared to signal-based paradigms. It is important to note that the performance gap between the proposed method and RL-based baselines does not solely stem from algorithmic superiority, but from a fundamental difference in control abstraction. RL-based traffic signal controllers, including MP-Light, CoLight, and AttentionLight, operate at the phase level and rely on aggregated traffic states, which inherently limits their ability to resolve fine-grained vehicle interactions under high saturation. Under extremely high traffic density, several learning-based signal control baselines exhibit even worse performance than fixed-time traffic lights. This phenomenon is mainly due to the limitation of phase-based decision making, where frequent signal switching and inter-agent coordination overhead introduce additional delay under saturated traffic conditions \cite{colight}. In contrast, traditional traffic lights maintain stable throughput by allocating long green phases, despite lacking adaptivity. The proposed method avoids this limitation by directly performing multi-agent trajectory-level motion planning, enabling continuous and fine-grained coordination among vehicles. As a result, it remains effective even under maximum traffic density where signal-based control paradigms tend to degrade.
By comparison, the proposed signal-free framework directly optimizes continuous vehicle trajectories, enabling explicit conflict resolution at the vehicle level.
Therefore, the comparison in this work should be interpreted as a paradigm-level evaluation between phase-based signal control and trajectory-level intersection management, which is consistent with the objective of assessing the necessity of traffic signals in future CAV-dominated environments.

The real-world inference performance of the proposed framework is verified on a computing platform equipped with an Intel Core i7-12650H CPU, NVIDIA GeForce RTX 4060 GPU, and 64 GB of RAM. For the 10,000-step simulation, path planning is triggered every 4 steps (0.4 s), resulting in a total of 2500 planning queries. The framework achieves an average inference latency of 7.75 ms per planning query, which is substantially lower than the 400-ms planning refresh interval used in the simulation, satisfying the real-time update requirements of the simulated urban intersection scenarios. This computational efficiency enables deployment on edge computing units (e.g., roadside units or intersection controllers), facilitating a centralized-distributed hybrid architecture that is compatible with existing V2X communication standards. Such deployability is critical for bridging the gap between theoretical planning algorithms and practical intelligent transportation systems.

\section{Conclusion}

This paper presents DSIP, a diffusion-based framework for signal-free intersection management. Comprehensive evaluations in SUMO demonstrated that trajectory-level coordination can significantly outperform traditional signalized control and state-of-the-art reinforcement-learning-based signal controllers in terms of delay, average speed, and safety, particularly under high-saturation conditions. Notably, the framework achieves an average inference latency of 7.75 ms per planning query, which is substantially lower than the 400-ms planning refresh interval used in the simulation. 

While this study focused on isolated four-leg intersections under idealized CAV conditions, this setup was deliberately chosen to establish a performance baseline for the signal-free paradigm, consistent with foundational studies \cite{dresner2004}. By isolating the coordination logic from mixed-traffic complexities, we validated the theoretical upper bound of trajectory-level management. Future work will extend this framework from three core dimensions: first, we will incorporate realistic communication delays and control-tracking deviations into the simulation framework; second, we will extend the current isolated intersection scenario to network-level coordinated control across multiple intersections, and adapt the framework to more complex traffic geometries such as roundabouts; third, we will relax the fixed line-following constraint at the intersection center to further explore the planning freedom of the diffusion model, and maximize the traffic efficiency potential of the signal-free paradigm. Additional research directions include direct comparisons with representative reservation-based and optimization-based AIM methods under a unified simulation interface.

Advancements in V2X communication and autonomous driving technologies continue to bridge the gap between simulation and deployment. This work provides empirical evidence supporting the viability of diffusion-based multi-agent coordination as a sustainable alternative to conventional signal control for future intelligent transportation systems. Ultimately, this research contributes to the vision of next-generation intelligent transportation systems, where intersection efficiency is defined by collaborative intelligence rather than fixed infrastructure, paving the way for safer, greener, and more scalable urban mobility.

\bibliographystyle{IEEEtran} 
\bibliography{refs}

\begin{IEEEbiography}[{\includegraphics[width=1in,height=1.25in,clip,keepaspectratio]{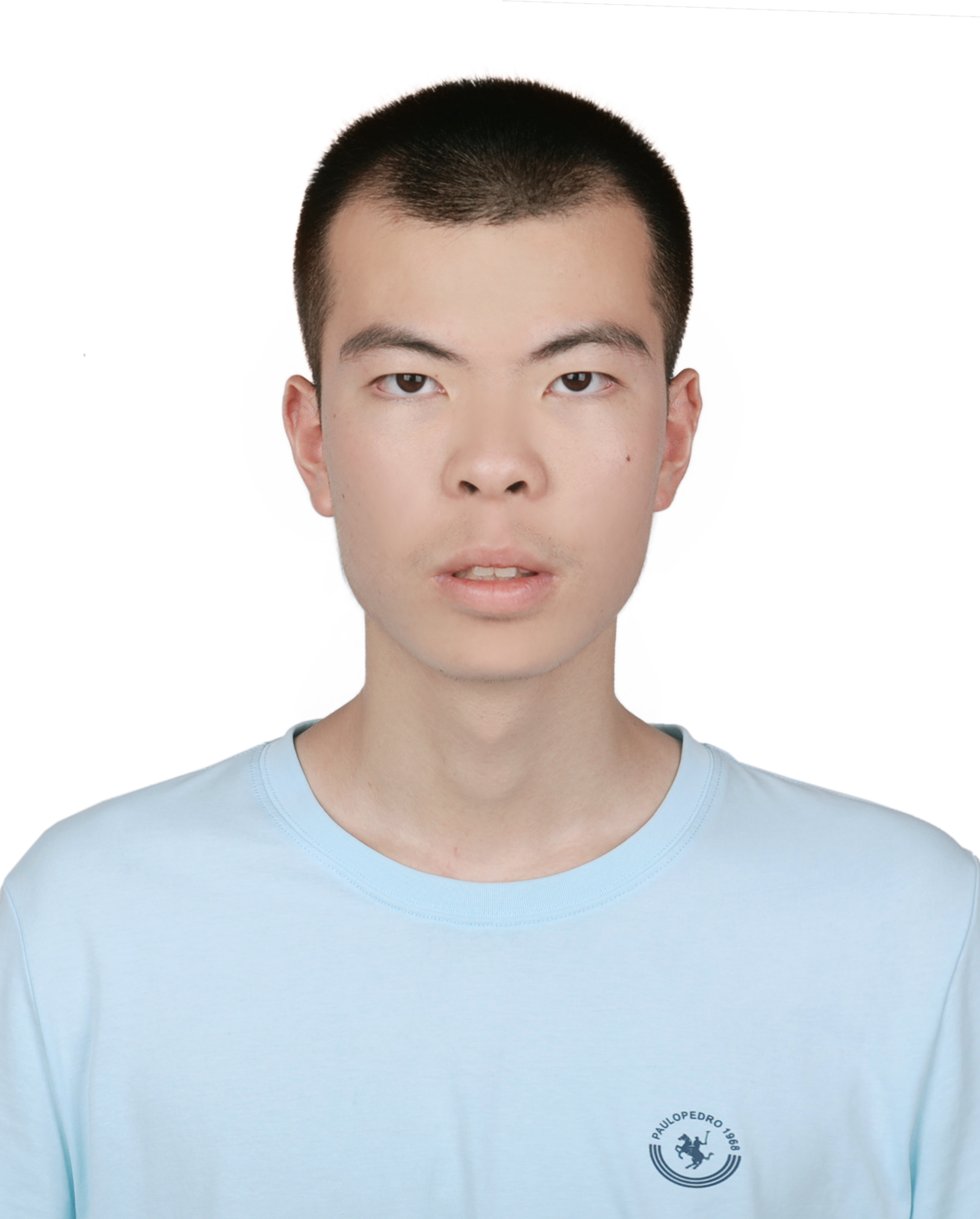}}]{Qian Hu}
is a sophomore at the Global Institute of Future Technology (GIFT), Shanghai Jiao Tong University, Shanghai, China, where he is currently pursuing the B.S. degree in sustainable energy technology. His main research interests include deep learning and autonomous driving.
\end{IEEEbiography}

\begin{IEEEbiography}[{\includegraphics[width=1in,height=1.25in,clip,keepaspectratio]{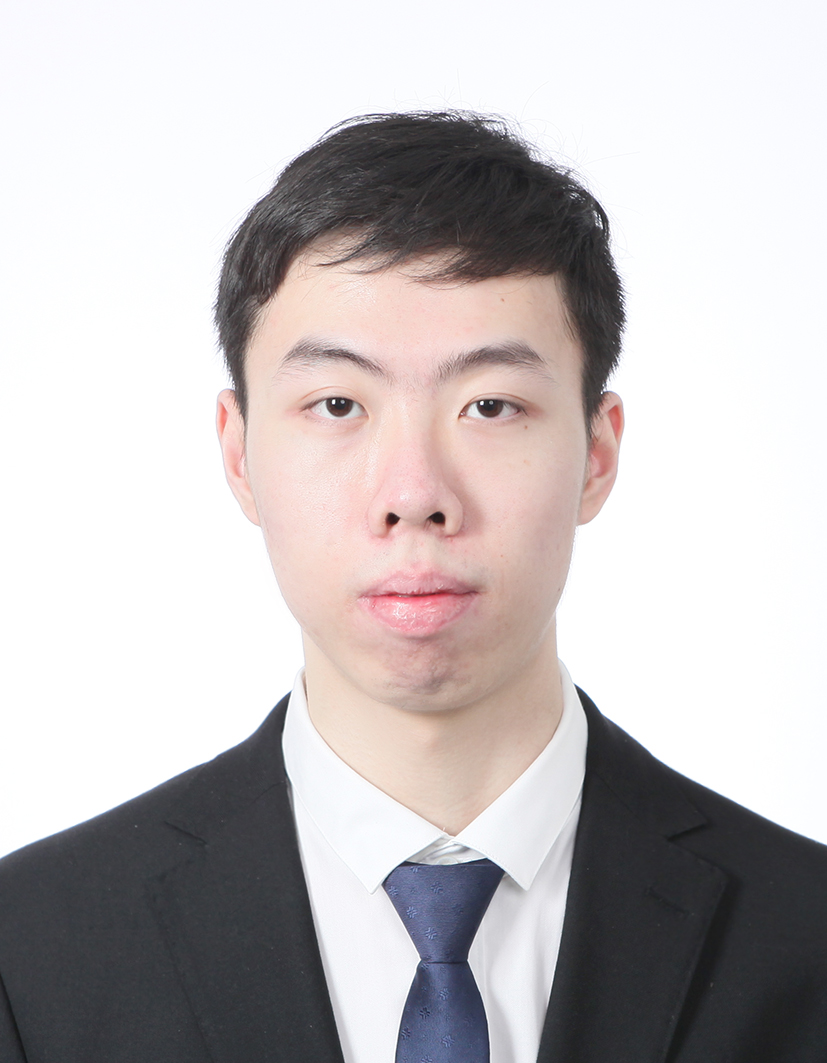}}]{Haoyang Peng}
received the B.S. degree in artificial intelligence from Shanghai Jiao Tong University, Shanghai, China, in 2025, and is currently pursuing the M.S. degree in electronic information engineering at the same university. His main research interests include deep learning and autonomous driving.
\end{IEEEbiography}

\begin{IEEEbiography}[{\includegraphics[width=1in,height=1.25in,clip,keepaspectratio]{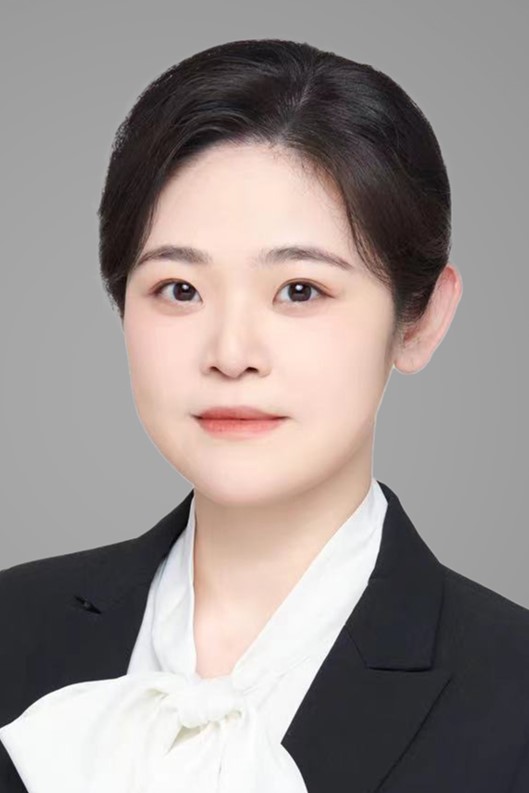}}]{Songan Zhang}
received the B.S. and M.S. degrees in automotive engineering from Tsinghua University, Beijing, China, in 2013 and 2016, respectively. In 2021, she earned the Ph.D. degree in mechanical engineering from the University of Michigan, Ann Arbor, MI, USA. Upon graduation, she joined Ford Motor Company as a Research Scientist, where she made contributions to pioneering innovations in smart manufacturing and advanced driver assist systems. She is currently an Assistant Professor with the Global Institute of Future Technology, Shanghai Jiao Tong University, Shanghai, China. Her research interests include accelerated and safety evaluation of autonomous vehicles; verification methods for autonomous driving systems; model-based, trustworthy, and data-efficient reinforcement learning and meta-learning; and the application of foundation models for decision-making in autonomous vehicles and intelligent transportation systems.
\end{IEEEbiography}

\begin{IEEEbiography}[{\includegraphics[width=1in,height=1.25in,clip,keepaspectratio]{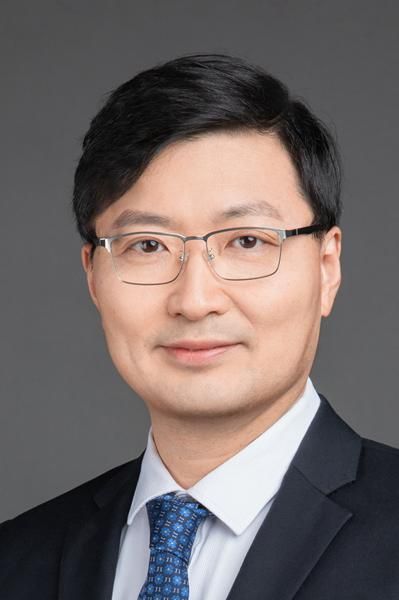}}]{Ming Yang}
received the M.S. and Ph.D. degrees from Tsinghua University, Beijing, China, in 1999 and 2003, respectively. He is currently a Full Tenure Professor at Shanghai Jiao Tong University and the Deputy Director of the Innovation Center of Intelligent Connected Vehicles. He has been working in the field of intelligent vehicles for more than 20 years. He participated in several related research projects, such as the THMR-V project (the first intelligent vehicle in China), European CyberCars and CyberMove projects, CyberC3 project, CyberCars-2 project, ITER transfer cask project, and AGV systems, etc.
\end{IEEEbiography}

\begin{IEEEbiography}
[{\includegraphics[width=1in,height=1.25in,clip,keepaspectratio]{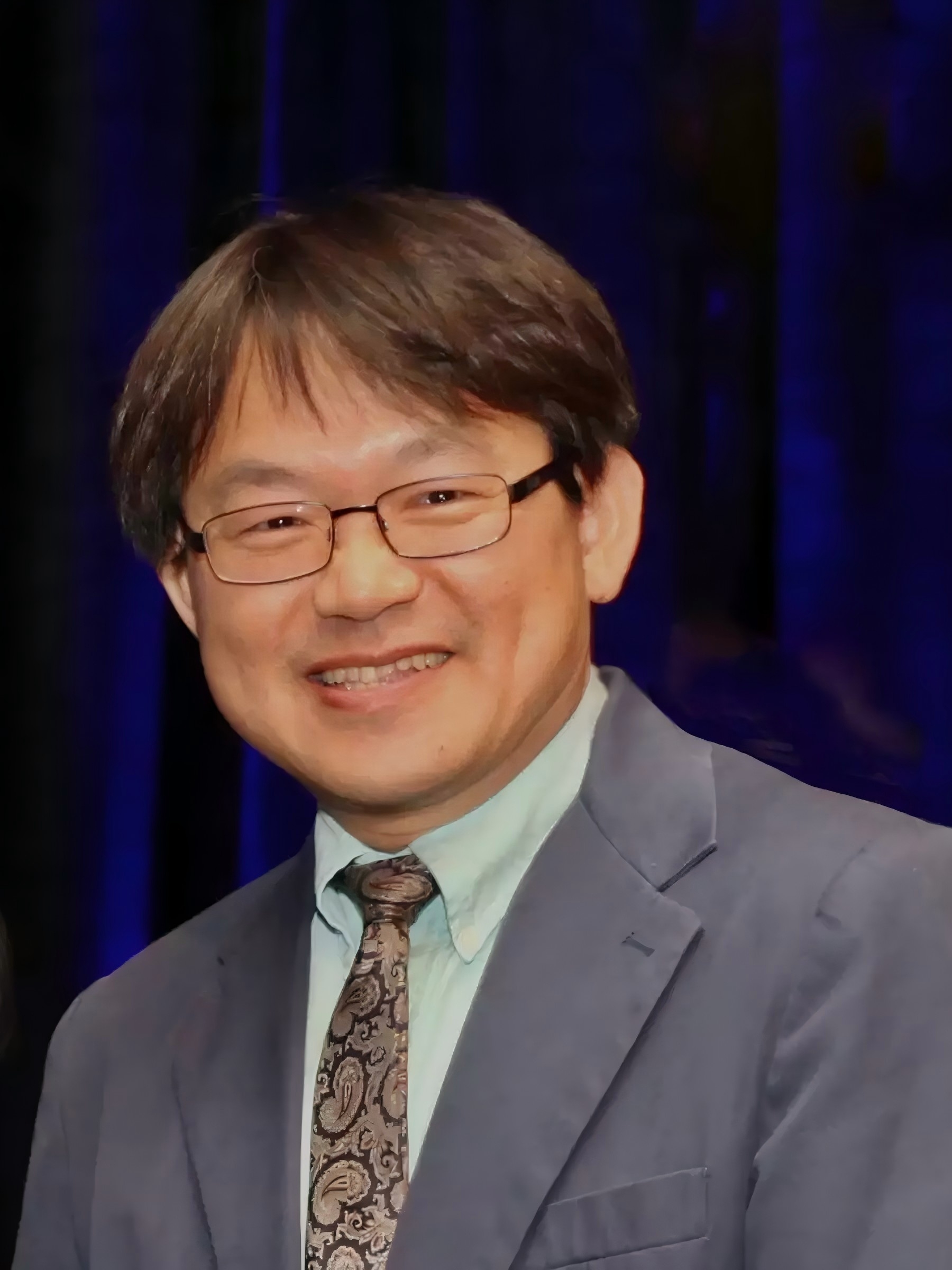}}]{Hongtei Eric Tseng}
(Member, IEEE) received the B.S. degree from National Taiwan University in 1986, and the M.S. and Ph.D. degrees in mechanical engineering from the University of California, Berkeley in 1991 and 1994, respectively. In May 2024, he joined as a Distinguished University Professor with the Department of Electrical Engineering, The University of Texas at Arlington. In 1994, he joined Ford Motor Company. At Ford from 1994 to 2022, he had a productive career and retired as a Senior Technical Leader of Controls and Automated Systems in research and advanced engineering. Many of his contributed technologies led to production vehicles implementation. He has over 100 U.S. patents and over 160 publications. He is a member of the National Academy of Engineering as of 2021. His technical achievements have been honored with the Ford’s Annual Technology Award, and the Henry Ford Technology Award, on seven occasions. Additionally, he was a recipient of the Control Engineering Practice Award from American Automatic Control Council in 2013, and a recipient of the Soichiro Honda Medal from the American Society of Mechanical Engineering in 2024.
\end{IEEEbiography}

\end{document}